\documentclass{article}

\usepackage{PRIMEarxiv}

\usepackage[utf8]{inputenc} 
\usepackage[T1]{fontenc}    
\usepackage{hyperref}       
\usepackage{url}            
\usepackage{booktabs}       
\usepackage{amsfonts}       
\usepackage{fancyhdr}       
\usepackage{graphicx}       

\usepackage{amsmath, amsfonts, amssymb, amsthm, dsfont, mathrsfs, hyperref, float, graphicx}

\newcommand*{\floor}[1]{ \left \lfloor #1 \right \rfloor }

\newtheorem{definition}{Definition}

\newtheorem{lemma}{Lemma}
\newtheorem{assumption}{Assumption}
\newtheorem{theorem}{Theorem}

\title{Homogenization of Multi-Agent Learning Dynamics in Finite-State Markov Games}

\author{
  Yann Kerzreho \\
  École Normale Supérieure de Paris-Saclay \\
  91190 Gif-sur-Yvette, France\\
  \texttt{yann.kerzreho@ens-paris-saclay.fr} \\
}
\date{Jun 2025}

\begin{document}

\maketitle

\begin{abstract}
This paper introduces a new approach for approximating the learning dynamics of multiple reinforcement learning (RL) agents interacting in a finite-state Markov game. The idea is to rescale the learning process by simultaneously reducing the learning rate and increasing the update frequency, effectively treating the agent's parameters as a slow-evolving variable influenced by the fast-mixing game state. Under mild assumptions—ergodicity of the state process and continuity of the updates—we prove the convergence of this rescaled process to an ordinary differential equation (ODE). This ODE provides a tractable, deterministic approximation of the agent's learning dynamics. An implementation of the framework is available at : https://github.com/yannKerzreho/MarkovGameApproximation
\end{abstract}


\section{Introduction}

\subsection{Background and Motivation} The study of multi-agent reinforcement learning (MARL) interacting over time in structured settings, such as Markov games (\cite{littman1994markov}), is a challenging task due to the high variance and non-stationarity of the learning process. Recently, an emergent literature faces this problem when trying to tackle the issue of algorithmic collusion \cite{schwalbe2018algorithms,calvano2020artificial,banchio2022artificial}. With the rise of algorithmic pricing on online market places, scientists wander if these algorithms might learn to cooperate on price and behave non-competitively. The first paper to introduce analytical tools in addition to intensive simulations was \cite{banchio2022artificial}. However, they applied their method to a stateless games, which is a huge simplification as RL methods are created to solve Markov decision process which include states. 

In their framework, the parameters defining the agents' behavior evolve according to a stochastic process whose updates depend on chosen actions. By rescaling time (increasing the update rate by a factor \(N\)) and reducing the update size accordingly (dividing the update magnitude by \(N\)), this approach, also known as hydrodynamic scaling, yields convergence to a deterministic ODE (\cite{kurtz1970solutions}). An accessible introduction to this fluid limit technique for Markov chains is given in \cite{darling2002fluid}. However, applying this method to games with state dependence is non-trivial. The state evolution, being discrete and rapidly fluctuating, is not suited for fluid approximations. One common workaround is to replace the state with a probability distribution on states, by introducing a population structure representation (see \cite{michaelides2019geometric,darling2008}). Then we could update the agent's parameters with an average update across states weighted by the distribution. A population structure would be useful for approximating agents playing many different games simultaneously and in parallel. However, our case is different as we want to approximate what happen for a single game.

Moreover, this work differs from the two-scale stochastic approximation of \cite{borkar1997stochastic} as our aim is to provide a means of analyzing the agent's learning period, even if the agents do not converge to a fixed strategy. The work of \cite{allmeier2024computing} is closely related, but does not apply to our case because the assumptions made are different: since they assume the existence of an exponentially stable attractor for the ODE, their dynamics cannot explode. In MARL, such an assumption cannot be made, as there is no guarantee of convergence to a fixed point for the traditional algorithm. In this way, they use stochastic approximation where we use homogenization techniques.

\subsection{Contributions}
The paper propose a novel homogenization-based method for approximating the learning dynamics of RL agents in a finite-state Markov game. Specifically, we model the agent’s parameters as a slow variable and the game’s state as a fast-mixing variable. By simultaneously scaling down the learning rate and scaling up the update frequency, we show that the stochastic process describing the agent’s behavior converges to an ODE. 

This ODE describes the evolution of the agent’s parameters averaged over the stationary distribution of the fast state variable. This type of result is known from multi-scale homogenization (\textit{e.g.} \cite{pavliotis2008multiscale}), though we have not found any existing frameworks that apply directly to our case. Therefore, we provide a new proof and framework tailored to our context.

To our knowledge, this approximation is original and offers new analytical tools for studying algorithmic behavior in dynamic environments. In particular, it has promising applications in the study of algorithmic collusion, where simulation-heavy methods currently dominate. Our framework allows for robust and reproducible analysis using ODEs, facilitating the design and testing of novel learning strategies.

\subsection{Structure}

In section \ref{sec2}, we will present the modeling framework for a reinforcement agent playing a finite-state Markov game and how we can derive a Markov chain from it. Section \ref{sec3} presents the assumptions and the main convergence theorem. Section \ref{sec4} showcases how Q-tables equipped with a right policy can be used in our framework. Finally, Section \ref{sec5} discusses the approximation and its usage.

\section{Framework} \label{sec2}
\subsection{Definitions}
Let us start by defining a Markov game, even if this object is often used, its notation and formal definition varies across the literature. Intuitively, a finite‐state Markov game is just a multi‐player Markov decision process where everyone’s action jointly determines the next state.

\begin{definition}[Markov Game] \label{def:markovGame}
    A finite state and action Markov game is a tuple $\mathcal G = (\mathcal I,\mathcal S, \nu_0, (\mathcal A^i)_{i \in \mathcal I}, T, (R^i)_{i \in \mathcal I})$ where,
    \begin{itemize}
        \item $\mathcal I = \{1, \dots, N\}$ is the set of $N$ players,
        \item $\mathcal S = \{1,\dots, S\}$ is the finite state space,
        \item $\nu_0$ is the initial distribution of state of the game,
        \item $\mathcal A^i = \{1, \dots, A^i\}$ is the finite set of player $i$'s actions, with $\mathcal A = \times _{i\in \mathcal I}\mathcal A^i$ the joint actions space,
        \item $T : \mathcal S \times \mathcal A \rightarrow \Delta(\mathcal S)$ is the state transition kernel, assigning a probability distribution over next states ($\Delta(\mathcal S)$) given a current state and joint action. Therefore, $T(s,a)(s')$ is the probability to jump at state $s'$ from the state $s$ when the joint action $a$ is played,
        \item $R^i : \mathcal S \times \mathcal A \rightarrow \mathbb R$ is a reward function of player $i$.
    \end{itemize}
\end{definition}

At each iteration $k$ of the game, each agent chooses an action $a^i_k \in \mathcal A^i$ according to the current state $s_k \in \mathcal S$, then the game transitions to state $s_{k+1}$ according to $T$, and each agent $i$ is rewarded by $r^i_k = R^i(s_k, a_k)$. Extending the formalism of \cite{banchio2022artificial}, we define a \textit{reinforcer}, an algorithm (e.g., a Q-table) that can play a Markov game, as follow:

\begin{definition}[Reinforcer] \label{def:reinforcer}
    A reinforcer indexed by $i$ is a tuple $(X^i_0, f^i, \pi^i)$ with,
\begin{itemize}
    \item A initial vector of parameters $X_0^i$,
    \item An update function $f : \mathcal A \times \mathbb R \times \mathbb R^{d_i} \times \mathcal S \times \mathcal S \rightarrow \mathbb R^{d_i}$ such that,
    \[X^i_{k+1} = X^i_k + f^i\left(a_k, r^i_k, X^i_k, s_k, s_{k+1}\right),\]
    \item A policy $\pi^i : \mathbb{R}^{d_i} \times \mathcal S \rightarrow \Delta(\mathcal A^i)$ that maps $X^i$ and state to the probability of playing each action.
\end{itemize}
\end{definition}

I have $a_k \in \mathcal A$ the action played by players, $r^i_k$ the reward of player $i$, and $s_k \in \mathcal S$ the state of the game, all at iteration $k$. When all players are reinforcers, we can aggregate them all in a single $d$-dimensional vector, $d = \sum_{i \in \mathcal I} d_i$, using a single update function $f$ and a single policy $\pi$. We create a new object to simplify the notation of reinforcers playing a Markov game by “wrapping” each period’s state‐action‐next‐state tuple into a single enlarged state. We will see that the entire learning process (parameters plus environment) becomes a single Markov chain with such definition.

\begin{definition} [Wrapped Markov Game] \label{def:WMG}
    Let a list of reinforcers (Def. \ref{def:reinforcer}) noted $RL = (X^i_0, f^i, \pi^i)_{i\in \mathcal I}$ playing a finite state Markov game (Def. \ref{def:markovGame}) noted $\mathcal G = (\mathcal I,\mathcal S, \nu_0, (\mathcal A^i)_{i \in \mathcal I}, T, (R^i)_{i \in \mathcal I})$. The wrapped Markov game associated is $\mathcal W(RL, \mathcal G) = (f,P, X_0, \mu_0)$. We define $E = \mathcal S \times \mathcal A \times \mathcal S$ the wrapped-state space and $\mathcal M$ the space of stochastic matrix of dimension $\#E$.
    \begin{itemize}
        \item The aggregate update of the parameters is $f : \mathbb R ^d \times E \rightarrow \mathbb R^d$ such that,
            \[
            f(x, g) = \bigoplus_{i=1}^n f_i(x_i, g) = \begin{pmatrix}
            f_1(x_1, g) \\
            \vdots \\
            f_n(x_n, g)
            \end{pmatrix}
            \quad \text{with } x = (x_1, \dots, x_n),\ x_i \in \mathbb{R}^{d_i}
            \]  
        \item A parametrized transition matrix for the wrapped-state $P : \mathbb R^d \rightarrow \mathcal M$, such that for $x \in \mathbb R^d$ and $g = (s_c,a,s_n),g'=(s_c',a',s_n') \in E$,
            \[
            P_x(g,g') = \mathds 1 \{s_n = s_c' \}  \; \pi\big(x+ f(x, g),s_c'\big)(a') \;  T\big(s_c',a' \big)(s_n'),
            \]
        where $\pi(x, s) = \bigotimes_{i=1}^n \pi_i(x_i, s)$. The element $s_c$ is the current state, $a$ is the joint action played and $s_n$ is the next state, which is also needed to update the parameters. The probability $\pi\big(x+ f(x, g),s_c'\big)(a')$ is the one of the learners picking joint action $a'$ in the state $s_c'$, after updating parameters by $f$.
        \item The initial aggregated parameters $X_0 = (X_0^1, \dots, X_0^N)$ ,
        \item The initial distribution of the wrapped game on $E$ noted $\mu_0$, such that for $g = (s_c,a,s_n) \in E$,
            \[
            \mu_0(g) = \nu_0(s_c) \; \pi\big(X_0, s_c \big)(a) \; T\big(s_c, a \big)(s_n).
            \]
    \end{itemize}
\end{definition}

The wrapped game is a way to simplify the processes produced by playing a Markov game (Def. \ref{def:markovGame}). To update a reinforcer (Def. \ref{def:reinforcer}) we need the current state of the game, the joint action chosen by players and the next state. All this information is summarize by an element of the wrapped-state space $E$. Moreover, updating the wrapped-state is also very simple as it uses a single parametrized transition matrix $P_x$. Now, to define the trajectories of such wrapped game, we introduce a probability space $(\Omega, \mathcal F, \mathbb P)$.

\begin{definition} [Wrapped Trajectories] \label{def:trajectories}
    The wrapped trajectories of a wrapped Markov game (Def. \ref{def:WMG}) is map,
    \[
    \big(X_n, G_n\big)_{n \in \mathbb N} : \Omega \rightarrow \big(\mathbb R^d, E)^{\mathbb N}
    \]
    such that for all $x\in \mathbb R^d$ and $g,g' \in E$,
    \begin{equation} \label{eq:trajectories}
    \begin{split}
        & X_{n+1} = X_n +  f(X_n, G_n), \\
        & \begin{split} P_{x}(g,g') & = \mathbb P\big(G_{n+1} = g' |G_n=g, X_{n+1} = x +f(x, g)\big) \\ & = \mathbb P\big(G_{n+1} = g' |G_n=g, X_n=x\big). \end{split}
    \end{split}
\end{equation}
We see that $G_n = \big(s_n, a_n, s_{n+1} \big)$ where $s_n$ and $s_{n+1}$ are the states of the game at iteration $n$ and $n+1$, while $a_n$ the joint actions played at iteration $n$. The joint actions $a_n$ is chosen with the state $s_n$ and the parameters $X_n$.  
\end{definition}

From the definition, we see that the chain $\big(X_n, G_n\big)_{n \in \mathbb N}$ is Markovian, but time‐inhomogeneous. This motivated the definition of the wrapped game. On $(\Omega,\mathcal{F},\mathbb{P})$, the process $(X_n,G_n)_{n\in\mathbb{N}}$ is the Markov chain of initial law $\delta_{X_0}\otimes\mu_0$ and transition kernel $(x,g)\;\longmapsto\;\delta_{\,x + f(x,g)}\;\otimes\;P_x\bigl(g,\cdot\bigr)$.

\subsection{Scaled process and Result Aimed}
Our approach is to divide by $N$ the update of the reinforcers and multiply by $N$ the number of iteration. Thus, we are going to show the convergence of the trajectories of a wrapped game scaled by $N$ towards a deterministic ODE. Let us define $(X^N_n, G^N_n)_{n \in \mathbb N}$ the trajectories when we changed the reinforcers to $(X^i_0, N^{-1}f^i, \pi^i)_{i\in \mathcal I}$. Then we have for all $x\in \mathbb R^d$ and $g,g' \in E$,
\begin{equation} \label{eq:scaledTrajectories}
    \begin{split}
        & X^N_{n+1} = X^N_n +  N^{-1}f(X_n, G_n), \\
        & \begin{split} P^N_{x}(g,g') & = \mathbb P\big(G^N_{n+1} = g' |G^N_n=g, X^N_{n+1} = x + N^{-1}f(x, g)\big) \\ & = \mathbb P\big(G^N_{n+1} = g' |G^N_n=g, X_n=x\big). \end{split}
    \end{split}
\end{equation}

Then, we are going to show the convergence of $X_{\lfloor Nt \rfloor}^N$ to the ODE defined by $ y(0)=X_0$ and $y'(t)=\beta(y(t))$. The limit derivative $\beta$ is defined as,
\begin{equation} \label{eq:fluidlimit}
    \beta = \lim_{N \rightarrow \infty} \beta^N, \quad \beta^N : x \in \mathbb R^d \mapsto \sum_{g \in E} \mu^N_x (g) f(x,g) \in \mathbb R^d,
\end{equation}
where $\mu^N_x$ is the ergodic probability measure of the chain given by $P^N_x$ under an assumption of uniform ergodicity in $x$ and $N$.

\subsection{Random Induction and change in Probability}

This subsection is purely technical and can be avoided. To facilitate the analysis, we build an explicit expression for the wrapped game transitions. A function $h^N : \mathbb R^d \times E \times [0,1] \rightarrow E$ can be built such that $G^N_{n+1} = h^N(X^N_n, G^N_n, \xi_{n+1})$ where $(\xi_n)_{n \in \mathbb N}$ are  \textit{i.i.d.} uniformly distributed on $[0,1]$. This defines a process with the same distribution as the original Markov chain, but with an explicit sampling structure. Moreover $h$ can be written,
\begin{equation*}
h^N(x, g, \xi) = \begin{cases}
\displaystyle 1 \quad \text{if } \xi \in \left[0, P^N_x(g, 1) \right] = I^N_{x,g}(1) \\[1mm]
\displaystyle i \quad \text{if } \xi \in \left]\sum_{j=1}^{i-1} P^N_x(g, j), \ \sum_{j=1}^{i} P^N_x(g, j) \right] = I^N_{x,g}(i), \text{ for } i > 1,
\end{cases}
\end{equation*} 
where $\bigsqcup_{i=1}^{\#E} I_{x,g}^N(i) = [0,1]$ and with $\lambda$ the Lebesgue measure, $\lambda(I_{x,g}^N(i)) = P^N_x(g, i)$. The intervals $I_{x,g}^N(i)$ are defining the inverse CDFs used for sampling transitions. To simplify notation, we will keep them unchanged despite this important underlying change. Therefore, if we prove the convergence of this new Markov chain (with changed probability) towards an ODE with $N \rightarrow \infty$, this will only prove the converge in distribution between the initial Markov chain (with unchanged probability) and the ODE.

\section{Theoretical Results} \label{sec3}

\subsection{Set up and Assumptions} Let $(\Omega, \mathcal F, \mathbb P)$ be a probability space, $\mathcal G = (\mathcal I,\mathcal S, (\mathcal A^i)_{i \in I}, T, (R^i)_{i \in I})$ a Markov game (Def. \ref{def:markovGame}) played by $(X^i_0, N^{-1}f^i, \pi^i)_{i \in \mathcal I}$ a set of reinforcers (Def. \ref{def:reinforcer}) parametrized by $N$. The associated wrapped Markov Game (Def. \ref{def:WMG}) is $\mathcal W^N = (N^{-1}f, P^N, X_0, \mu^N_0)$ and the wrapped trajectories (Def. \ref{def:trajectories}) are $(X^N_n, G^N_n)_{n \in \mathbb N}$. From here we will make the following assumptions:
\begin{assumption} \label{ass:update}
    The update function $f:\mathbb R^d \times E \rightarrow \mathbb R^d$ is Lipschitz in its first argument. Moreover, there exist a compact $D \subsetneq \mathbb R^d$ such that for any $n,N \in \mathbb N$, $(X^N_n, G^N_n) \in D\times E$. Hence, $f$ is bounded on $D$ and we note $\sup_D f = \| f \|_{\infty}$.
\end{assumption}

\begin{assumption} \label{ass:policy}
    The joint policy $\pi:\mathbb R^d \times \mathcal S \rightarrow \Delta(\mathcal A)$ is $L_\pi$-Lipschitz in its first argument on D.
\end{assumption}

\begin{assumption} \label{ass:kernel}
    For any $x\in D$, the transition matrix $P^N_x$ respects uniformly the Doeblin condition: there exist an integer $k$, a positive $c$ and a probability measure $q$ on E such that $(P_x^N)^k(i,j) \ge c \ q(j)$ for any $x \in D$, any $N \in \mathbb N$, and any $i,j \in \mathcal S$. Hence the transition matrix $P^N_x$ is ergodic and we note $\mu^N_x$ the ergodic probability measure of the chain given by $P^N_x$.
\end{assumption}

We will see in Section \ref{sec4} that these assumptions are easy to verify for a Q table equipped with a continuous policy. Note that the existence of $D$ can be proven because the rewards are bounded. Whenever updates attempt to approximate the cumulative discounted rewards over the iterations of the game, such a compact space exists. The assumption \ref{ass:kernel} can be verified simply by taking a uniform exploration rate across actions.

\subsection{Main Results}
From these assumptions we can deduce the following lemmas:
\begin{lemma} \label{lemma1}
    Under the assumptions \ref{ass:update} and \ref{ass:policy}, $P^N_x$ is Lipschitz in its argument $x$. This mean that with the operator norm $\|\cdot\|_{\mathrm{op}} = \sup_{\nu \in \mathcal{P}(E)} \|\nu (\cdot)\|_{\mathrm{TV}}$ where $\mathcal P(E)$ is the set of probability distributions over $E$, there exist $L_P$ such that for any $x,y \in D$ and any $N \in \mathbb N$,
    \begin{equation*}
        \left\|P^N_x - P^N_y\right\|_{\mathrm{op}} \le L_P \|x - y\|.
    \end{equation*}
\end{lemma}

\begin{lemma} \label{lemma2}
    Under the assumptions \ref{ass:update} and \ref{ass:kernel}, the map $x \mapsto \mu^N_x$, where $\mu^N_x$ is the unique invariant measure of the Markov kernel $P^N_x$, is Lipschitz continuous. Hence $\beta^N$ and $\beta$ defined in (\ref{eq:fluidlimit}) are Lipschitz continuous on $D$.
\end{lemma}

\begin{lemma} \label{lemma3}
    Under the assumptions \ref{ass:update}, \ref{ass:policy} and \ref{ass:kernel}, $\| \beta^N(x) - \beta(x) \| = \mathcal O (N^{-1}).$
\end{lemma}

The proofs can be found in appendix \ref{proof:lemma1}, \ref{proof:lemma2} and \ref{proof:lemma3} respectively. Now we can pass to the main result of our paper:
\begin{theorem} \label{thm:disctApprox}
    Under Assumptions \ref{ass:update}, \ref{ass:policy} and \ref{ass:kernel}, the scaled process $X^N_{\lfloor Nt \rfloor}$ converges weakly to the solution of the ODE defined by $\dot y(t) = \beta(y(t))$ and $y(0) = X_0$, where $\beta$ is defined as,
    \[
    \beta = \sum_{g \in E} \mu_x (g) f(x,g).
    \]
\end{theorem}

\subsection{Proof of the Main Theorem}
\begin{proof}

Recall that we defined the Euler approximation by 
\[
y^N_0=X_0,\quad y^N_{n+1}=y^N_n+\frac{1}{N}\beta(y^N_n).
\]

By the triangular inequality, for any \(T>0\) we have
\[
\mathbb{E}\sup_{0\le t\le T}\|X^N_{tN} - y(t)\| \le \mathbb{E}\sup_{0\le t\le T}\|X^N_{tN} - y^N_{tN}\| + \sup_{0\le t\le T}\|y^N_{tN} - y(t)\|,
\]
and by standard results for the convergence of the Euler scheme, since $\beta$ is Lipschitz, we have
\[
\sup_{0\le t\le T}\|y^N_{tN} - y(t)\| = \mathcal O\left(\frac{1}{N}\right).
\]

Then, to prove the theorem, we just need to show that,
\[
\sup_{0\le t\le T}\|X^N_n - y^N_{tN}\| = \mathcal o\left(1\right).
\]

\noindent\textbf{Step 1.} We will start by dividing the sum of updates into block of $M \in \mathbb N$ (s.t. $tN/M \in \mathbb N$) updates and a residual that can be treated with a Grönwall lemma,
\begin{equation} \label{eq:proof:1} \begin{split} 
    \mathbb E \left\| X^N_{tN} - y^N_{tN} \right\| & = \frac{1}{N}  \mathbb E \left\| \sum_{n=0}^{tN-1} f(X_n^N, G_n^N) -   \beta(y^N_n) \right\| \\
    & \le \frac{1}{N} \mathbb E  \left\| \sum_{n=0}^{tN-1} f(X_n^N, G_n^N) - \beta(X_n^N) \right\| + \frac{1}{N} \mathbb E  \left\| \sum_{n=0}^{tN-1} \beta(X_n^N) -  \beta(y^N_n) \right\|\\
    & \le  \frac{1}{N} \sum_{k=0}^{M-1} \mathbb E  \left\| \sum_{n=ktN/M}^{(k+1)tN/M - 1} f(X_n^N, G_n^N) - \beta(X_n^N) \right\| + \frac{1}{N} \mathbb E  \left\| \sum_{n=0}^{tN-1} \beta(X_n^N) -  \beta(y^N_n) \right\|.
\end{split} \end{equation}
We add and subtract $\beta(X_n^N)$ before using the triangular inequality to pass to the second line. Each term of the sum over $k$ can be decomposed as follows, for simplicity we change indices to $k=0$,
\begin{equation} \begin{split} \label{eq:proof:2}
    \mathbb E \left\| \sum_{n=0}^{tN/M-1} f(X_n^N, G_n^N) -  \beta(X_n^N) \right\| & \le \mathbb E \left\| \sum_{n=0}^{tN/M-1} f(X_n^N, G_n^N) -  f( X_n^N, \tilde G_n^N) \right\| \\
    &  + \sum_{n=0}^{tN/M-1} \mathbb E \left\| \beta(X_0^N)  - \beta(X_n^N) + f(X_n^N, \tilde G_n^N) -  f( X_0^N, \tilde G_n^N) \right\| \\
    & + \mathbb E \left\| \sum_{n=0}^{tN/M-1} f(X_0^N, \tilde G_n^N) - \beta^N(X_0^N) \right\| \\
    & + \frac{tN}{M} \left\| \beta^N(X_0^N) - \beta(X_0^N) \right\|,
\end{split} \end{equation}
where $(\tilde G_n^N)_{n \ge k}$ is the Markov chain defined by the kernel $P_{X^N_0}$ and $\tilde G_0^N = G_0^N$. Moreover $G_{n+1}^N = h(X_n^N, G_n^N, \xi_{n+1})$ and $\tilde G_{n+1}^N = h(X_0^N, \tilde G_n^N, \xi_{n+1})$. The chain $(\tilde G_n^N)_{n \in \mathbb N}$ is the homogeneous Markov chain of states when freezing the parameters to its first values.

\medskip\noindent\textbf{Step 2.} We bound $\mathbb E \left\| \sum_{n=0}^{tN/M-1} f(X_n^N, G_n^N) -  f( X_n^N, \tilde G_n^N) \right\|$ by using this inequality,

\begin{align*}
    & \mathbb E \left\| \sum_{n=0}^{tN/M-1} f(X_n^N, G_n^N) -  f( X_n^N, \tilde G_n^N) \right\| \le  \sum_{n=0}^{tN/M-1} \mathbb E \left\| \sum_{g \in E} f(X_n^N, g) \big( \mathds 1 \{ G_n^N = g\} - \mathds 1 \{\tilde G_n^N = g\} \big) \right\|  \\
    & \le \left\| f \right\|_{\infty} \sum_{n=0}^{tN/M-1} \mathbb E \left[ \sum_{g \in E} \left| \mathds 1 \{ G_n^N = g\} - \mathds 1 \{\tilde G_n^N = g\} \right| \right] \le \left\| f \right\|_{\infty} \sum_{n=0}^{tN/M-1} \mathbb P \left( G_n^N \ne \tilde G_n^N \right).
\end{align*}

Now, by the Lemma \ref{lemma4} described and proven bellow with $T=tN/M$, there exist a positive $C > 0$ such that,
\begin{align*}
    \left\| f \right\|_{\infty} \sum_{n=0}^{tN/M-1} \mathbb P \left( G_n^N \ne \tilde G_n^N \right) \le \left\| f \right\|_{\infty} \sum_{n=0}^{tN/M-1} 1 - {\left( 1 - C \frac{t}{M} \right)_+}^{\frac{tN}{M}} \le \left\| f \right\|_{\infty} \frac{tN}{M}\left( 1 - {\left( 1 - C \frac{t}{M} \right)_+}^{\frac{tN}{M}} \right).
\end{align*}  

\medskip\noindent\textbf{Step 3.} To bound $\mathbb E \sum_{n=0}^{tN/M-1} \left\| \beta(X_n^N)  - \beta(X_0^N) + f(X_n^N, \tilde G_n^N) -  f( X_0^N, \tilde G_n^N) \right\|$, we start by noticing that for any $n \in \mathbb N$,
\begin{equation*}
    \left\| X_{n+1}^N - X_n^N \right\| \leq \frac{1}{N} \| f \|_{\infty}
\end{equation*}
then we have,
\begin{equation*}
    \left\| X_L^N - X_0^N \right\| \le \sum_{l=0}^{L-1} \left\| X_{l+1}^N - X_l^N \right\| \leq \frac{L}{N} \| f \|_{\infty}
\end{equation*}
and finally with $L_\beta$ and $L_f$ the Lipschitz constant of $\beta$ and $f$,
\begin{align*}
    & \mathbb E \sum_{n=0}^{tN/M-1} \left\| \beta(X_0^N)  - \beta(X_n^N) + f(X_n^N, \tilde G_n^N) -  f( X_0^N, \tilde G_n^N) \right\|  \le (L_\beta + L_f) \mathbb E \sum_{n=0}^{tN/M-1} \left\| X_n^N - X_0^N \right\| \\
    & \le (L_\beta + L_f) \sum_{n=0}^{tN/M-1} \frac{n}{N} \| f \|_{\infty} \le (L_\beta + L_f) \frac{t^2N}{M^2} \| f \|_{\infty} = \mathcal O\left(\frac{t^2N}{M^2}\right).
\end{align*}  

\medskip\noindent\textbf{Step 4.} To bound $\mathbb E \left\| \sum_{n=0}^{tN/M-1} f(X_0^N, \tilde G_n^N) - \beta(X_0^N) \right\|$ we rewrite,

\begin{align*}
    & \mathbb E  \left\| \sum_{n=0}^{tN/M-1} f(X_0^N, \tilde G_n^N) -  \beta(X_0^N) \right\| \le \sum_{g \in E} \mathbb E  \left\| \sum_{n=0}^{tN/M-1} f(X_0^N, g) \mathds 1 \{\tilde G_n^N = g\} -  f(X_0^N,g)\mu_{X_0^N}(g) \right\| \\
    & \le \sum_{g \in E} \left\| f \right\|_{\infty} \mathbb E \left| \sum_{n=0}^{tN/M-1} \mathds 1 \{\tilde G_n^N = g\} -  \mu_{X_0^N}(g) \right| ,
\end{align*}
then by the Lemma \ref{lemma5} described bellow, if $tN/M \rightarrow \infty$ the order of the sum is $\left(\frac{tN}{M}\right)^{1/2}$.  

\medskip\noindent\textbf{Step 5.} In this step we aggregate step 2, 3 and 4 and inject them into the inequalities of step 1. We use the Lemma \ref{lemma3} to get a order of magnitude for the last term of the inequality \ref{eq:proof:2}. Then we have an order for inequality \ref{eq:proof:2} of:
\[
O\left( \frac{N}{M} \left( 1 - {\left( 1 -  \frac{1}{M} \right)_+}^{\frac{N}{M}}  \right) \right)  + \mathcal O\left(\frac{N}{M^2}\right) +  \mathcal O\left( \left( \frac{N}{M} \right)^{1/2} \right) + \mathcal O \left( \frac{1}{M}\right)
\]
Then we can use this order into the inequality \ref{eq:proof:1}. To get the order of first term, we multiply the previous order of inequality \ref{eq:proof:2} by $\frac{M}{N}$, this give an order of:
\[
O\left( 1 - {\left( 1 -  \frac{1}{M} \right)_+}^{\frac{N}{M}}  \right)  + \mathcal O\left(\frac{1}{M}\right) +  \mathcal O\left( \left( \frac{M}{N} \right)^{1/2} \right) + \mathcal O \left( \frac{1}{N} \right)
\]
By using $1 - (1-M^{-1})^{N/M} = 1 -\exp ( NM^{-1} \ln (1-M^{-1})) = 1 - \exp \big(NM^{-2} + o(NM^{-2})\big) \sim NM^{-2}$ when $N = o(M^{-2})$, we can find a $M$ optimizing the convergence rate. Let $M = N^\alpha$, then the order is:
\[
N^{1- 2\alpha} + N^{-\alpha} + N^{\alpha/2 - 1/2} + N^{-1}
\]
The minimum is for $\alpha = 3/5$, this gives a final order of $N^{-1/5}$. Therefore,
\begin{equation} \label{eq:proof:3}
    \mathbb E \left\| X^N_{tN} - y^N_{tN} \right\| \le  \mathcal O\left(N^{-1/5} \right) + \frac{1}{N} \mathbb E  \left\| \sum_{n=0}^{tN-1} \beta(X_n^N) -  \beta(y_n) \right\|
\end{equation}

\medskip\noindent\textbf{Step 6.} As the inequality \ref{eq:proof:3} is defined for any finite $t$, we can define $\kappa$, an error defined as,
\begin{equation*}
    \kappa(N) = \max_{n \in \{1, \dots, tN\}} \mathbb E \left\| X^N_{n} - y^N_{n} \right\| - \frac{1}{N} \mathbb E  \left\| \sum_{k=0}^{n-1} \beta(X_k^N) -  \beta(y^N_k) \right\|,
\end{equation*}
an application that allow to bound the error of order $N^{-1/5}$ independently of $n$. We denote $L_\beta$ the Lipschitz constant of $\beta$, then we have for $N$ large enough and for all $n \in \{0,\dots tN\}$, 
\begin{align*}
    & \mathbb E \left\| X^N_{n} - y^N_{n} \right\| 
    \le \kappa(N) + \frac{1}{N} \sum_{k=0}^{n-1} \mathbb E  \left\|  \beta(X_k^N) -  \beta(y_k^N) \right\| \\
    & \le \kappa(N) + L_\beta \frac{1}{N} \sum_{k=0}^{n-1} \mathbb E \left\| X_k^N - y_k^N \right\|
\end{align*}

By using this version of the Grönwall lemma: $u_n \le B + \sum_{k=0}^{n-1} \alpha_k u_k \implies u_n \le B\exp \left\{ \sum_{k=0}^{n-1} \alpha_k\right\}$, we have for all $n \in \{0,\dots tN\}$,
\begin{align*}
    \mathbb E \left\| X^N_{n} - y^N_{n} \right\| & \le \kappa(N) \exp\left\{ \sum_{k=0}^{n-1} L_\beta/N \right\} \\
    & \le  \kappa(N) e^{t L_\beta} = \mathcal O\left( N^{-1/5} \right)
\end{align*}

As we have the inequality for all $n \in \{0,\dots tN\}$ we pass to the sup to get the final result.
\end{proof}

\subsection{Additional Lemmas}
\begin{lemma} \label{lemma4}
    For $N \in \mathbb N$ and under the assumptions \ref{ass:update}, \ref{ass:policy} and \ref{ass:kernel}, with $(\xi_n)_{n \in \mathbb N} \sim \mathcal U[0,1]$ i.i.d. we define two Markov chains induced by the same sequence of random variable, $G_{n+1}^N = h^N(X_n^N, G_n^N, \xi_{n+1})$ et $\tilde G_{n+1}^N = h^N(X_0^N, \tilde G_n^N, \xi_{n+1})$. Then for any $T \in \mathbb N$, there is $C > 0$ such that for all $n \in \{0, ... T\}$
    \begin{equation*}
        \mathbb P \left( G_n^N \ne \tilde G_n^N \right) \le 1 - {\left( 1 - C \frac{T}{N} \right)_+}^T
    \end{equation*}
\end{lemma}

\begin{lemma} \label{lemma5}
    $(G_n)_{n \in \mathbb N}$ is a Markov chain in $(E, \mathcal P(E))$ with $E$ finite. Its transition matrix is denoted $P$ and $\mu^n$ is the distribution of $G_n$ for all $n$. If the chain respects the Doeblin condition: there exist an integer k, a positive c and a probability measure $q$ on E s.t. $P^k(i,j) \ge c \ q(j)$ then,  
    \begin{align*}
        & \|  \mu^n - \mu  \|_{TV} = \frac{1}{2} \sum_{g \in E} | \mu^n(g) - \mu(g)| \leq (1-c)^{\floor{n/k}} \\
        &  \mathbb E \left| \frac{1}{N} \sum_{n=1}^N \mathds 1 \{ G_n = g\} - \mu(g) \right| = \mathcal O(N^{-1/2}) \quad \text{for any } g \in E,
    \end{align*}
    where $\mu$ is the invariant probability of the Markov chain.
\end{lemma}

The proof of the Lemma \ref{lemma4} and the second point of Lemma \ref{lemma5} can be found in appendix (\ref{proof:lemma4}, \ref{proof:lemma5}). The first point of the Lemma \ref{lemma5} can be found in \cite{gine1997lectures}.

\section{Example: Q-table} \label{sec4}

\subsection{Description}

In this subsection, we present a classical reinforcer that satisfies the assumptions of the theorem. A Q-table assigns to each state-action pair $(a^i, s) \in \mathcal A^i \times \mathcal{S}$ a Q-value, which is updated using the Bellman equation to approximate the expected discounted rewards of taking action $a^i$ in state $s$. We denote by $X^i_k(a^i, s)$ the Q-value at iteration $k$ for agent $i$.

When Q-tables are paired with a policy, they can be interpreted as \textit{reinforcers}. The update rule is defined for all $(a^i, s)$ by:
\begin{equation*} \label{eq:Qtable}
    X^i_{k+1}(a^i, s) = 
    \begin{cases}
        X^i_k(a^i, s) + \alpha \left( r^i_k + \gamma \max\limits_{b \in A^i} X^i_k(b, s_{k+1}) - X^i_k(a^i, s) \right), & \text{if } (a^i, s) = (a^i_k, s_k), \\[2mm]
        X^i_k(a^i, s), & \text{otherwise,}
    \end{cases}
\end{equation*}
where $\gamma \in [0,1)$ is the discount factor and $\alpha$ the learning rate.

We define the policy $\pi^i$ as a softmax policy with temperature $\tau$, combined with uniform exploration at rate $\varepsilon$. For all $a^i \in \mathcal A^i$, we set:
\begin{equation*} \label{eq:softMax}
    \pi^i(X^i_k, s_k)(a^i) = 
    (1 - \varepsilon) \cdot \frac{\exp(X^i_k(a^i, s_k)/\tau)}{\sum\limits_{b \in \mathcal A^i} \exp(X^i_k(b, s_k)/\tau)} + \frac{\varepsilon}{\# \mathcal A^i}.
\end{equation*}

\subsection{Verification of the Assumptions} The verification of the set of assumptions will be briefly justified here. Assumption \ref{ass:update} holds thanks to the linear structure of the updates and noticing that the rewards are bounded. Note $\bar r$ the maximum absolute reward across the states, joint-actions and players. Then it holds from an induction that $X^i_k \in B(0, \|X_0\|_{\infty} \vee \bar r /(1 - \gamma))$ for any $k$. As the rewards are bounded, the discounted sum of rewards across iterations of the game are also bounded. For assumption \ref{ass:policy}, the policy $\pi$ is $L_\pi$-Lipsichtz on \textit{D} as its Jacobian is a continuous function, hence bounded on a compact.

The uniform Doeblin of assumption $\ref{ass:kernel}$ is fulfilled by to the $ \varepsilon$-rate of uniform exploration. In two iteration, the chain have a positive probability to explore any state of the wrapped game. Let $\kappa$ the smallest probability in the state-transition function $T$. Then, for any $x \in D$ and $s \in \mathcal S$, it holds that $\pi(x,s) \ge \varepsilon/\# \mathcal S$ and for all $i,j \in \mathcal S$, we have that $({P_x^N})^2(i,j) \ge \left( \kappa  \varepsilon/\# \mathcal S \right)^2$.

\section{Discussion} \label{sec5}

The two-scale homogenization technique presented in this paper provides a meaningful and rigorous approximation of the original process dynamics. Approximating the learning process by an ordinary differential equation (ODE) offers both advantages and limitations, which we now discuss.

On the positive side, deriving the closed-form ODE enables further analytical insights and allows simulating the learning dynamics without repeatedly running the full stochastic game. This can significantly reduce computational cost and facilitate large-scale experimentation and parameter exploration. However, one challenge lies in computing the limit $\beta$, which requires determining the stationary distribution $\mu_x$ at each time step. For large state spaces, this may necessitate numerical methods, potentially offsetting some of the computational gains.

Another aspect is that our approximation is insensitive to proportional rescalings of the learning rate across reinforcement learning algorithms. While this may be seen as an advantage—removing the influence of a sensitive hyperparameter—it could also be a limitation, as learning rates often play a critical role in the early phases of training.

The method is also highly flexible: the definition of the wrapped game can be extended to include, for instance, batched updates or stochastic rewards, allowing a broader class of reinforcement learners to fit within the framework.

Finally, we note a fundamental limitation: the ODE approximation is valid over finite time horizons, but its long-term accuracy depends on the stability of the dynamics. If the ODE exhibits attractor sets (e.g., fixed points or limit cycles), trajectories would remain close to the true dynamics; otherwise, divergence may occur.

\section*{Acknowledgments}
I would like to express my gratitude to Pierre Cardaliaguet and Yannick Viossat for supervising this work, which I completed during my first year of master's studies at Paris-Dauphine University. I would also like to warmly thank Aldric Labarthe for all his helpful discussions.

\bibliographystyle{unsrt}  
\bibliography{references}  

\appendix
\section{Additional Proofs}

\subsection{Lemma \ref{lemma1}.} \label{proof:lemma1}

    Under the assumptions \ref{ass:update} and \ref{ass:policy}, $P^N_x$ is Lipschitz in its argument $x$. This mean that with the operator norm $\|\cdot\|_{\mathrm{op}} = \sup_{\nu \in \mathcal{P}(E)} \|\nu (\cdot)\|_{\mathrm{TV}}$ where $\mathcal P(E)$ is the set of probability distributions over $E$, there exist $L_P$ such that for any $x,y \in D$ and any $N \in \mathbb N$,
    \begin{equation*}
        \left\|P^N_x - P^N_y\right\|_{\mathrm{op}} \le L_P \|x - y\|.
    \end{equation*}

\begin{proof}
    By decomposing the operator norm we have,
    \[ \begin{split}
    \| P^N_x - P_y^N \|_{op} &  = \sup_{\nu \in \mathcal P(E)} \| \nu P^N_x - \nu P_y^N \|_{TV} \\
    & = \sup_{\nu \in \mathcal P(E)} \frac{1}{2}\sum_{g'\in E} \sum_{g \in E} \nu(g) \left[  P^N_x(g,g') -  P^N_y(g,g') \right].
    \end{split} \]
    We just need to bound the difference between the two probability, by noticing that,
    \[ \begin{split}
    P^N_x(g,g') & = \mathbb P\big( G^N_{n+1} = g' | X^N_{n+1} = x + N^{-1}f(x, g), G^N_n = g\big) \\
    & = \mathds 1 \{s_n = s_c' \}  \; \pi\big(x+ N^{-1}f(x, g),s_c'\big)(a') \;  T\big(s_c',a' \big)(s_n').
    \end{split}\]
    we have with the policy $\pi$ being $L_\pi$-Lipsichtz on D and $f$ being bounded on D,
    \begin{align*}
        \big| P^N_x(g,g') - P^N_y(g,g') \big| & \le T\big(s_c',a' \big)(s_n') \big| \pi\big(x+ N^{-1}f(x, g),s_c'\big)(a') - \pi\big(y+ N^{-1}f(y, g),s_c'\big)(a')  \big| \\
        & \le  L_\pi \big| x+ N^{-1}f(x, i) -  y - N^{-1}f(y, i) \big| \\
        & \le L_\pi(1+L_f/N) \big\|x-y\big\|
    \end{align*}
    This proves the $ (\#E)^2L_\pi(1+L_f)$-Lipschitz continuity of the transition matrices for all $N$ on $D$.
\end{proof}

\subsection{Lemma \ref{lemma2}.} \label{proof:lemma2}

    Under the assumptions \ref{ass:update} and \ref{ass:kernel}, the map $x \mapsto \mu^N_x$ where $\mu^N_x$ is the unique invariant measure of the Markov kernel $P^N_x$, is Lipschitz continuous. Hence $\beta^N$ and $\beta$ defined in (\ref{eq:fluidlimit}) are Lipschitz continuous on $D$.

\begin{proof}
Let $x, y \in D$. Since $E$ is finite and $P^N_x$ satisfies a uniform Doeblin condition, there exists an integer $k \ge 1$ and a constant $\theta \in (0,1)$ such that for all probability measures $\mu, \nu \in \mathcal P(E)$,
\[
    \|\mu (P^N_x)^k - \nu (P^N_x)^k\|_{\mathrm{TV}} \le \theta \|\mu - \nu\|_{\mathrm{TV}}.
\]
To simplify the notation in this proof, we omit the exponent $N$, therefore $P_x^k$ mean $(P_x^N)^k$. Since $\mu_x P_x = \mu_x$, it holds that
\begin{equation*}
    \|\mu_x - \mu_y\|_{\mathrm{TV}} = \|\mu_x P_x^k - \mu_y P_y^k\|_{\mathrm{TV}}  \le \|\mu_x P_x^k - \mu_x P_y^k\|_{\mathrm{TV}} + \|\mu_x P_y^k - \mu_y P_y^k\|_{\mathrm{TV}}.
\end{equation*}

Since the second term can be bounded with the contraction property, we still have to bound the first term $\|\mu_x P_x^k - \mu_x P_y^k\|_{\mathrm{TV}}$. Let us prove by induction that for all $k \in \mathbb N$,
\begin{equation*}
    \|P_x^k - P_y^k\|_{\mathrm{op}} \le k L_P \|x - y\|,
\end{equation*}
where $\|\cdot\|_{\mathrm{op}} = \sup_{\mu \in \mathcal{P}(E)} \|\mu (\cdot)\|_{\mathrm{TV}}$ and $L_P$ is the Lipschitz constant of $x \mapsto P_x$ on $D$ such that,
\begin{equation*}
\|P_x - P_y\|_{\mathrm{op}} \le L_P \|x - y\|.
\end{equation*}

Assume the property holds for some $k \ge 1$. Then,
\begin{align*}
    \|P_x^{k+1} - P_y^{k+1}\|_{\mathrm{op}} 
    &= \|P_x P_x^k - P_y P_y^k\|_{\mathrm{op}} \le  \|P_x (P_x^k - P_y^k)\|_{\mathrm{op}} + \|(P_x - P_y) P_y^k\|_{\mathrm{op}} \\
    &\le \|P_x\|_{\mathrm{op}} \cdot \|P_x^k - P_y^k\|_{\mathrm{op}} + \|P_x - P_y\|_{\mathrm{op}} \cdot \|P_y^k\|_{\mathrm{op}} \\
    &\le k L_P \|x - y\| + L_P \|x - y\| = (k+1)L_P \|x - y\|.
\end{align*}

Therefore, using $\mu_x \in \mathcal{P}(E)$,
\begin{equation*}
    \|\mu_x P_x^k - \mu_x P_y^k\|_{\mathrm{TV}} \le \|P_x^k - P_y^k\|_{\mathrm{op}} \le k L_P \|x - y\|.
\end{equation*}

Use the contraction property of $P_y^k$,
\begin{equation*}
\|\mu_x - \mu_y\|_{\mathrm{TV}} \le k L_P \|x - y\| + \theta \|\mu_x - \mu_y\|_{\mathrm{TV}}.
\end{equation*}

Which gives,
\begin{equation*}
    \|\mu_x - \mu_y\|_{\mathrm{TV}} \le \frac{k L_P}{1 - \theta} \|x - y\|.
\end{equation*}
\end{proof}

\subsection{Lemma \ref{lemma3}.} \label{proof:lemma3}

    Under the assumptions \ref{ass:update}, \ref{ass:policy} and \ref{ass:kernel}, $\| \beta^N(x) - \beta(x) \| = \mathcal O (N^{-1}).$

\begin{proof}
    We start from the definition,
    \[
    \| \beta^N(x) - \beta(x) \| \le \left\| \sum_{g \in E} f(x,g) \big(\mu^N_x (g) - \mu_x (g)\big) \right\| \le 2 \| f\|_{\infty} \|\mu^N_x - \mu_x\|_{TV},
    \]
    then from classical perturbation theory for finite state Markov chain we have $C > 0$ such that,
    \[
    \| \mu^N_x - \mu_x \|_{TV} \le C \| P^N_x - P_x^{\infty} \|_{op}.
    \]
    As the proof of the Lemma \ref{lemma1},
    \[ 
    \| P^N_x - P_x^{\infty} \|_{op}  = \sup_{\nu \in \mathcal P(E)} \frac{1}{2}\sum_{g'\in E} \sum_{g \in E} \nu(g) \left[  P^N_x(g,g') -  P^{\infty}_x(g,g') \right].
    \]
    and,
    \[
    P^N_x(g,g') = \mathds 1 \{s_n = s_c' \}  \; \pi\big(x+ N^{-1}f(x, g),s_c'\big)(a') \;  T\big(s_c',a' \big)(s_n').
    \]
    Then we have,
    \[\begin{split}
    \left|  P^N_x(g,g') -  P^{\infty}_x(g,g') \right| & \le T\big(s_c',a' \big)(s_n') \; \big| \pi\big(x+ N^{-1}f(x, g),s_c'\big)(a') - \pi(x,s_c')(a') \big| \\
    & \le T\big(s_c',a' \big)(s_n') \frac{L_\pi \| f \|_{\infty}}{N},
    \end{split}\]
    and by summing over $g$ and $g'\in E$ we can conclude,
    \[
    \| \mu^N_x - \mu_x \|_{TV} = \mathcal O \left( \| P^N_x - P_x^{\infty} \|_{op} \right) = \mathcal O \left( N^{-1}\right)
    \]
\end{proof}

\subsection{Lemma \ref{lemma4}.} \label{proof:lemma4}

    For $N \in \mathbb N$ and under the assumptions \ref{ass:update}, \ref{ass:policy} and \ref{ass:kernel}, with $(\xi_n)_{n \in \mathbb N} \sim \mathcal U[0,1]$ i.i.d. we define two Markov chains induced by the same sequence of random variable, $G_{n+1}^N = h^N(X_n^N, G_n^N, \xi_{n+1})$ et $\tilde G_{n+1}^N = h^N(X_0^N, \tilde G_n^N, \xi_{n+1})$. Then for any $T \in \mathbb N$, there is $C > 0$ such that for all $n \in \{0, ... T\}$
    \begin{equation*}
        \mathbb P \left( G_n^N \ne \tilde G_n^N \right) \le 1 - {\left( 1 - C \frac{T}{N} \right)_+}^T
    \end{equation*}

For ease the reading of the following proof, we rewrite the induction map $h^N$,\begin{equation*}
h^N(x, g, \xi) = \begin{cases}
\displaystyle 1 \quad \text{if } \xi \in \left[0, P^N_x(g, 1) \right] = I^N_{x,g}(1) \\[1mm]
\displaystyle i \quad \text{if } \xi \in \left]\sum_{j=1}^{i-1} P^N_x(g, j), \ \sum_{j=1}^{i} P^N_x(g, j) \right] = I^N_{x,g}(i), \text{ for } i > 1.
\end{cases}
\end{equation*}

\begin{proof}
We start by using the Bayes formula to write,
\begin{align*}
    \mathbb P \left(G_{n+1}^N = \tilde G_{n+1}^N \right) \ge \mathbb P \left( G_n^N = \tilde G_n^N \right) \mathbb P\left(G_{n+1}^N = \tilde G_{n+1}^N \ | \ G_n^N = \tilde G_n^N \right)
\end{align*}

We are looking for a lower bound on $\mathbb P\left(G_{n+1}^N = \tilde G_{n+1}^N \ | \ G_n^N = \tilde G_n^N \right)$ to induce an upper bound of $\mathbb P\left(G_{n+1}^N \ne \tilde G_{n+1}^N \right)$. Remember that
\begin{equation*}
    \mathbb P (G^N_{n+1} = g | G_n^N, X_n^N) = P_{X_n^N}\big(g, G_n^N\big).
\end{equation*}

Let us recall that the intervals $I_{x,g}^N(i)$ are defining the inverse CDFs used for the sampling transitions of $G^N_{n+1}$ when $G^N_n = g$ and $X^N_n = x$. Now, let us see that if we have $G_{n}^N = \tilde G_{n}^N$, then we have $G_{n+1}^N = \tilde G_{n+1}^N$ if and only if there exist a $i \in \{1, \dots \#E\}$ such that the random variable $\xi_{n+1}$ is in both $I^N_{X^N_n, G^N_n}(i)$ and $I^N_{X^N_0, \tilde G^N_n}(i)$. At each step, the two chains will remain equals if the random variable $\xi$ lands in a common region of their respective transition partitions. Then we can write,
\begin{align*}
    \mathbb P\left(G_{n+1}^N = \tilde G_{n+1}^N \ | \ G_n^N = \tilde G_n^N \right) & = \lambda \left( \bigcup_{i=1}^{\#E} \left( I^N_{X^N_n, G^N_n}(i) \ \cap \ I^N_{X^N_0, \tilde G^N_n}(i) \right) \right) \\
    & = \sum_i^{\#E} \lambda \left( I^N_{X^N_n, G^N_n}(i) \ \cap \ I^N_{X^N_0, \tilde G^N_n}(i) \right) .
\end{align*}

As we need to control the intersection of two segment, let's look at the difference of their upper and lower bounds. Whenever $G^N_n = \tilde G^N_n = g$, with $P^N$ being $L_P$-Lipschitz and $\| X^N_m - X^N_n \| \le \frac{|m-n|}{N} \| f \|_{\infty}$, it holds for any $i >1$ that,
\begin{equation*}
    \begin{split}
        \big| \min I^N_{X^N_n, G^N_n}(i) - \min I^N_{X^N_0, \tilde G^N_n}(i) \big| & \le  \sum_{j=1}^{i-1} \big| P^N_{X^N_n}(g, j) - P^N_{X^N_0}(g, j) \big| \\
        & \le L_P \sum_{j=1}^{i-1} \big| X^N_n - X^N_0 \big| \le L_P (i-1) \frac{n}{N} \| f \|_{\infty},
    \end{split}
\end{equation*}
and for $i<\#E$ that,
\begin{equation*}
    \begin{split}
        \big| \max I^N_{X^N_n, G^N_n}(i) - \max I^N_{X^N_0, \tilde G^N_n}(i) \big| & \le  \sum_{j=1}^{i} \big| P^N_{X^N_n}(g, j) - P^N_{X^N_0}(g, j) \big| \\
        & \le L_P \sum_{j=1}^{i} \big| X^N_n - X^N_0 \big| \le L_P i \frac{n}{N} \| f \|_{\infty},
    \end{split}
\end{equation*}

We also know that for any two intervals: $I_1 = [a_1,\,b_1], I_2 = [a_2,\,b_2]$,
\[
\lambda(I_1 \cap I_2) = \bigl(\min\{b_1,b_2\} - \max\{a_1,a_2\}\bigr)_+.
\]
Using our notation we have,
\[
\lambda\bigl(I^N_{X^N_n,g}(i) \cap I^N_{X^N_0,g}(i)\bigr)
 \ge \Bigl(
\min \bigl\{\max I^N_{X^N_n,g}(i),\,\max I^N_{X^N_0,g}(i)\bigr\}
-
\max\bigl\{\min I^N_{X^N_n,g}(i),\,\min I^N_{X^N_0,g}(i)\bigr\}
\Bigr)_+.
\]
Therefore using our previous bounds,
\begin{align*}
\lambda\bigl(I^N_{X^N_n,g}(i)\cap I^N_{X^N_0,g}(i)\bigr)
& \ge 
\Bigl(\bigl[\max I^N_{X^N_0,g}(i) - L_P\,i\,\tfrac{n}{N}\lVert f\rVert_\infty\bigr]
 - 
\bigl[\min I^N_{X^N_0,g}(i) + L_P\,(i-1)\,\tfrac{n}{N}\lVert f\rVert_\infty\bigr]
\Bigr)_+\\
& = 
\Bigl(\underbrace{\bigl[\max I^N_{X^N_0,g}(i) - \min I^N_{X^N_0,g}(i)\bigr]}_{=\,P^N_{X^N_0}(g,i)}
 - L_P\,(i-1)\,\tfrac{n}{N}\lVert f\rVert_\infty 
 - L_P\,i\,\tfrac{n}{N}\lVert f\rVert_\infty
\Bigr)_+\\
& = 
\Bigl(P^N_{X^N_0}(g,i)  -  L_P\,(2\,i - 1)\,\tfrac{n}{N}\,\lVert f\rVert_\infty\Bigr)_+.
\end{align*}

And finally, with $\sum_{i=1}^{\#E} P^N_{X^N_0}(g,i) = 1$, $\sum_{i=1}^{\#E} (2i -1) \le 2\#E$, and $n \le T$, there is a $C>0$ such that,
\begin{align*}
    & \mathbb P\left(G_{n+1}^N = \tilde G_{n+1}^N \ | \ G_n^N = \tilde G_n^N \right) = \sum_{i=1}^{\#E} \lambda \left( I^N_{X^N_n, G^N_n}(i) \ \cap \ I^N_{X^N_0, \tilde G^N_n}(i) \right) \\
    & \ge \left( 1 - 2 L_P \| f \|_{\infty} \frac{T}{N} \#E \right)_+ = \left( 1 - C \frac{T}{N} \right)_+.
\end{align*}

With this lower bound and $G_0^N = \tilde G_0^N$, an induction can be used to get,
\begin{align*}
    & \mathbb P \left(G_{n+1}^N = \tilde G_{n+1}^N \right) \ge \left( 1 - C \frac{T}{N} \right)_+ \mathbb P \left( G_n^N = \tilde G_n^N \right)  \\
    & \mathbb P \left( G_n^N = \tilde G_n^N \right) \ge {\left( 1 - C \frac{T}{N} \right)_+}^n \ge {\left( 1 - C \frac{T}{N} \right)_+}^T.
\end{align*}
\end{proof}

\subsection{Lemma \ref{lemma5}.} \label{proof:lemma5}

    $(G_n)_{n \in \mathbb N}$ is a Markov chain in $(E, \mathcal P(E))$ with $E$ finite. Its transition matrix is denoted $P$ and $\mu^n$ is the distribution of $G_n$ for all $n$. If the chain respects the Doeblin condition: there exist an integer k, a positive c and a probability measure $q$ on E s.t. $P^k(i,j) \ge c \ q(j)$ then,  
    \begin{align*}
        & \|  \mu^n - \mu  \|_{TV} = \frac{1}{2} \sum_{g \in E} | \mu^n(g) - \mu(g)| \leq (1-c)^{\floor{n/k}} \\
        &  \mathbb E \left| \frac{1}{N} \sum_{n=1}^N \mathds 1 \{ G_n = g\} - \mu(g) \right| = \mathcal O(N^{-1/2}) \quad \text{for any } g \in E,
    \end{align*}
    where $\mu$ is the invariant probability of the Markov chain.

\begin{proof}
The first point is classical results, we refer to \cite{gine1997lectures} first section. Set 
\[
\alpha_\ell  = \sup_{i\in E}\bigl\|P^\ell(i,\cdot)-\mu(\cdot)\bigr\|_{TV}
 \le (1-c)^{\lfloor \ell/k\rfloor},
\]
so in particular, by recalling the definition of the total variation,
\(\bigl|\mathbb P(G_n=g)-\mu(g)\bigr|\le 2\alpha_n\)
and also,
\(\bigl|\mathbb P(G_j=g\mid G_i=g)-\mu(g)\bigr|\le 2\alpha_{j-i}\). Let \(S_N(g)=\sum_{n=1}^N\mathds 1\{G_n=g\}.\)

The bias therm is,
\[
\Bigl|\mathbb E[S_N(g)/N]-\mu(g)\Bigr|
 = \frac1N\sum_{n=1}^N\bigl|\mathbb P(G_n=g)-\mu(g)\bigr|
 \le \frac{2}{N}\sum_{n=1}^N\alpha_n
 = O\bigl(N^{-1}\bigr).
\]

The variance term is,
\[
\text{Var}(S_N(g))
=\sum_{n=1}^N\text{Var}(\mathds 1\{G_n=g\})
 + 2\sum_{1\le i<j\le N}\text{Cov}\bigl(\mathds 1\{G_i=g\},\mathds 1\{G_j=g\}\bigr).
\]
We have \(\text{Var}(\mathds 1\{G_n=g\})\le\tfrac14\), and moreover,
\(\bigl|\text{Cov}\bigl(\mathds 1\{G_i=g\},\mathds 1\{G_j=g\}\bigr)\bigr|\le 2\alpha_{j-i} + 2\alpha_j,\) as,
\[ \begin{split}
    \text{Cov}\bigl(\mathds{1}\{G_i=g\},\,\mathds{1}\{G_j=g\}\bigr)
     & = 
    \mathbb P(G_i=g)\,\bigl[\mathbb P(G_j=g\mid G_i=g)-\mathbb P(G_j=g)\bigr] \\
     &\le 
    \bigl|\mathbb P(G_j=g\mid G_i=g)-\mu(g)\bigr|
     + \bigl|\mathbb P(G_j=g)-\mu(g)\bigr|.
\end{split} \]

Then we have,
\[
\text{Var}(S_N(g))
 \le \frac N4  + 4 \sum_{h=1}^{N-1}(N-h)\ \alpha_h + 4 \sum_{j=2}^{N-1} (j-1) \ \alpha_j
 = O(N),
\]

hence by Jensen (or Cauchy–Schwarz),
\[
\mathbb E\Bigl|S_N(g)/N - \mathbb E[S_N(g)]/N\Bigr|
 \le \frac{\sqrt{\text{Var}(S_N(g))}}N
 = O\bigl(N^{-1/2}\bigr).
\]

By combining the bias and the Jensen bound we conclude,
\[
\mathbb E\Bigl|\tfrac1N S_N(g)-\mu(g)\Bigr|
 = O\bigl(N^{-1/2}\bigr).
\]
\end{proof}

\end{document}